\documentclass[10pt,twocolumn,letterpaper]{article}

\usepackage[pagenumbers]{cvpr} 

\definecolor{cvprblue}{rgb}{0.21,0.49,0.74}
\usepackage[pagebackref,breaklinks,colorlinks,allcolors=cvprblue]{hyperref}

\def\paperID{16298} 
\def\confName{CVPR}
\def\confYear{2025}

\title{CoSDH: Communication-Efficient Collaborative Perception via Supply-Demand Awareness and Intermediate-Late Hybridization}

\author{
Junhao Xu\textsuperscript{\rm 1}\footnotemark[1],
Yanan Zhang\textsuperscript{\rm 2}\thanks{Equal Contribution. $^{\dag}$Corresponding Author.},
Zhi Cai\textsuperscript{\rm 1},
Di Huang\textsuperscript{\rm 1$\dag$}\\
\textsuperscript{\rm 1}State Key Laboratory of Complex and Critical Software Environment, Beihang University, Beijing, China\\
\textsuperscript{\rm 2}School of Computer Science and Information Engineering, Hefei University of Technology, Hefei, China\\
{\tt\small junhaoxu@buaa.edu.cn, yananzhang@hfut.edu.cn, \{caizhi97, dhuang\}@buaa.edu.cn}
}

\newcommand{\mymethodname}{\texttt{CoSDH}}

\begin{document}
\maketitle
\begin{abstract}
Multi-agent collaborative perception enhances perceptual capabilities by utilizing information from multiple agents and is considered a fundamental solution to the problem of weak single-vehicle perception in autonomous driving. However, existing collaborative perception methods face a dilemma between communication efficiency and perception accuracy. To address this issue, we propose a novel communication-efficient collaborative perception framework based on supply-demand awareness and intermediate-late hybridization, dubbed as \mymethodname. By modeling the supply-demand relationship between agents, the framework refines the selection of collaboration regions, reducing unnecessary communication cost while maintaining accuracy. In addition, we innovatively introduce the intermediate-late hybrid collaboration mode, where late-stage collaboration compensates for the performance degradation in collaborative perception under low communication bandwidth. Extensive experiments on multiple datasets, including both simulated and real-world scenarios, demonstrate that \mymethodname~ achieves state-of-the-art detection accuracy and optimal bandwidth trade-offs, delivering superior detection precision under real communication bandwidths, thus proving its effectiveness and practical applicability. The
code will be released at \href{https://github.com/Xu2729/CoSDH}{https://github.com/Xu2729/CoSDH}.
\end{abstract}
\section{Introduction}
\label{sec:introduction}

Collaborative perception allows multiple agents to exchange complementary perception information. It fundamentally addresses the issues of limited perception range, sensor blind spots, and occlusion from obstacles inherent in single-agent perception, thereby improving both the range and accuracy of perception. Recent studies have demonstrated that collaborative perception can be applied to various autonomous driving tasks, including 3D object detection~\cite{chen2019f, xu2022opv2v, xu2022v2x}, semantic segmentation~\cite{li2022v2x, xu2023cobevt}, 3D occupancy prediction~\cite{song2024collaborative}, and trajectory prediction~\cite{yu2023v2x}, exhibiting enhanced performance and improved robustness to occlusions, thus making it a crucial approach for advancing autonomous driving systems.

\begin{figure*}
  \centering
  \begin{subfigure}{0.365\linewidth}
    \includegraphics[width=\textwidth]{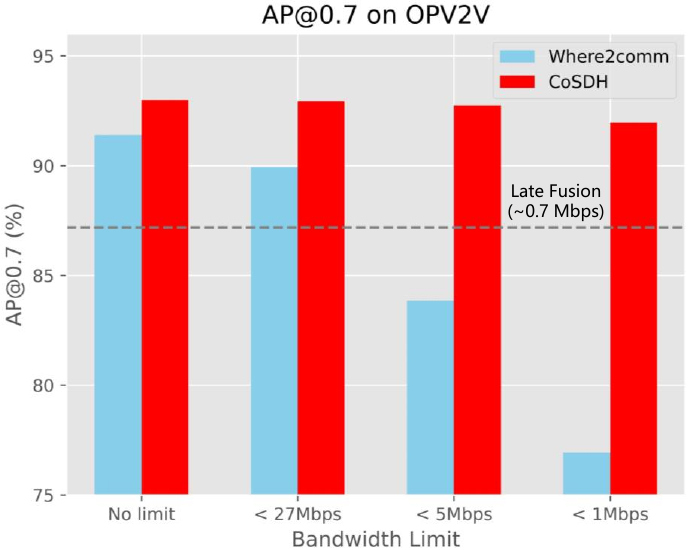}
    \caption{The 3D detection accuracy of Where2comm~\cite{hu2022where2comm} and \mymethodname~ on OPV2V dataset~\cite{xu2022opv2v} with different bandwidth limit. Assume the number of collaborative agents is 4 and detection frequency is 10Hz.}
    \vspace{-5pt}
    \label{fig:intr-b}
  \end{subfigure}
  \hfill
  \begin{subfigure}{0.615\linewidth}
    \includegraphics[width=\textwidth]{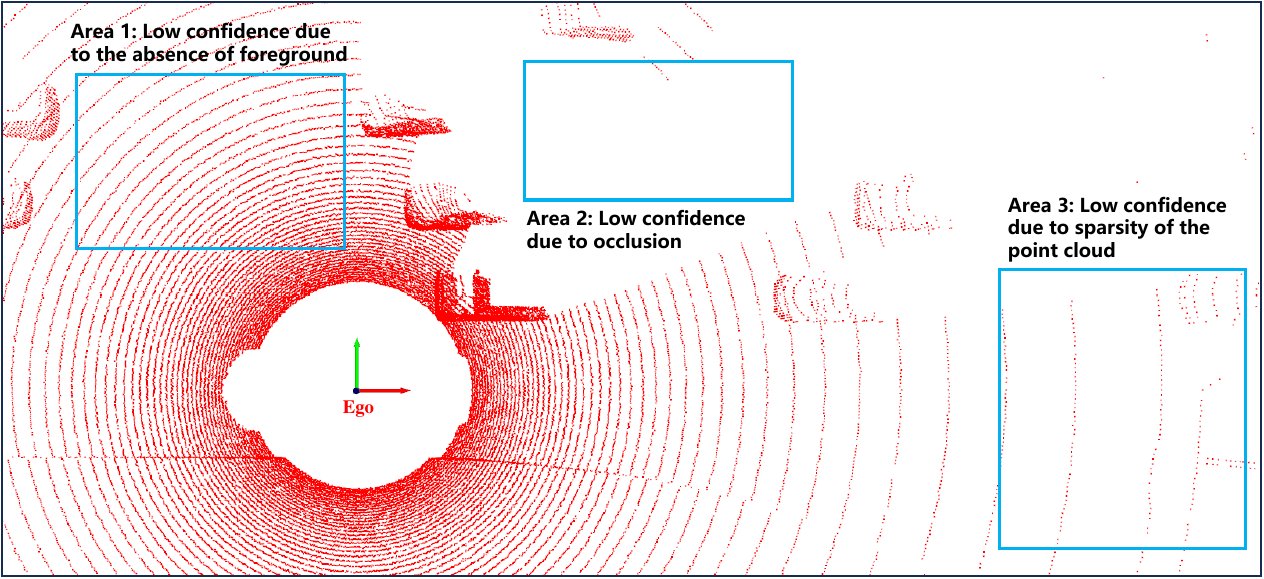}
    \caption{Point cloud map and classification of areas with low foreground confidence. Although Area 1 has low confidence due to the absence of foreground, it allows for good observation without collaboration. In contrast, Areas 2 and 3 exhibit poor observation and require collaboration, with Area 2 having low confidence due to occlusion and Area 3 due to sparsity of the point cloud.}
    \vspace{-5pt}
    \label{fig:intr-a}
  \end{subfigure}
  \caption{Issues of existing communication-efficient collaborative perception methods.}
  \vspace{-12pt}
  \label{fig:intr}
\end{figure*}

Communication efficiency is a key issue in collaborative perception. An early approach to improving communication efficiency was to use autoencoders to compress the intermediate features that need to be transmitted~\cite{xu2022opv2v}. Later, communication-efficient methods such as Where2comm~\cite{hu2022where2comm} were proposed, which select important and sparse foreground regions for collaboration. However, these methods still lack sufficient precision in area selection and require more bandwidth than the real-world constraints\footnote{To the best of our knowledge, the common V2X communication approach uses IEEE 802.11p-based DSRC (Dedicated Short-Range Communications) technology~\cite{arena2019overview, lozano2019review}, which provides an information transmission rate of approximately 27 Mbps.}.  As shown in Fig.~\ref{fig:intr-b}, when the bandwidth is limited to 27 Mbps, Where2comm~\cite{hu2022where2comm} experiences a decrease of about 1.5\% in average precision (AP@0.7) on the OPV2V~\cite{xu2022opv2v} dataset with an intersection-over-union (IoU) threshold of 0.7. When the bandwidth is further limited to 5 Mbps and 1 Mbps, AP@0.7 drops by approximately 7.5\% and 14.5\%, respectively, even falling below the accuracy of late fusion methods at the result level. This makes it difficult for existing collaborative perception methods to meet the demands of practical applications.

In this paper, we first analyze the problem of selecting collaboration areas and refine this selection based on the supply-demand relationship between agents. As shown in Fig.~\ref{fig:intr-a}, for Ego, low-foreground confidence areas can be divided into three types: Area 1 (low conficence due to absence of foreground), Area 2 (low conficence due to occlusion), and Area 3 (low conficence due to sparsity of the point cloud). Although all three areas appear as background from the Ego's perspective, Area 1 is well-observed and does not require collaboration, while Areas 2 and 3, with poorer observation, need collaboration. Additionally, to address the issue of significant accuracy degradation under real bandwidth constraints, we propose that compensating intermediate collaboration results with late fusion is a feasible solution. As shown in Fig.~\ref{fig:intr-b}, late fusion achieves much higher accuracy than Where2comm~\cite{hu2022where2comm} even with a bandwidth of approximately 0.7 Mbps, compared to Where2comm's accuracy with bandwidth less than 5 Mbps. So the use of intermediate-late hybrid collaboration can greatly improve the accuracy lower bound under low-bandwidth conditions.

Based on these insights, we propose \mymethodname, a novel communication-efficient collaborative perception framework for 3D object detection. As shown in Fig.~\ref{fig:framework}, \mymethodname~ consists of three key components: i) Supply-demand-aware information selection, which chooses sparse yet crucial regions for collaboration; ii) Intermediate feature transmission and fusion, which transmits and effectively aggregates information from multiple agents in a communication-efficient manner; iii) Confidence-aware late fusion, which compensates for the intermediate fusion results at a minimal communication cost to improve accuracy. To evaluate \mymethodname, we conducted extensive experiments on the simulation datasets OPV2V~\cite{xu2022opv2v}, V2XSim~\cite{li2022v2x}, and the real-world dataset DAIR-V2X~\cite{yu2022dair}. The experimental results show that i) In the baseline case, \mymethodname~ uses less bandwidth while achieving satisfactory accuracy, demonstrating a better accuracy-bandwidth trade-off; ii) Under the simulated real-world communication condition with a total bandwidth limit of 27 Mbps, \mymethodname~ experiences less accuracy degradation and outperforms other methods while using less bandwidth.

In summary, our contributions are as follows:
\begin{itemize}
\item We present \mymethodname, an innovative communication-efficient collaborative perception framework with better accuracy-bandwidth trade-offs for 3D object detection.
\item We propose a novel supply-demand-aware information selection module, further refining the collaboration area selection to achieve more efficient communication.
\item We design a novel intermediate-late hybrid collaborative perception paradigm, where confidence-aware late fusion compensates for the intermediate fusion results to maintain high accuracy under low-bandwidth conditions.
\item We conduct extensive experiments across multiple datasets with bandwidth constraints set closer to real-world conditions, and \mymethodname~ achieves state-of-the-art detection accuracy along with optimal bandwidth trade-offs, demonstrating its feasibility in real-world collaborative perception scenarios.
\end{itemize}

\section{Related Work}
\label{sec:related_work}

\subsection{3D Object Detection}

3D object detection, a key technology in autonomous driving, identifies and localizes objects in 3D scenes using environmental data from onboard sensors and can be categorized into image-based, point-cloud-based, and multimodal-fusion-based methods~\cite{qian20223d, wang2023multi, ma2024vision, mao20233d}. Image-based methods can be further classified into monocular~\cite{wang2019pseudo,chen2016monocular,ku2019monocular}, stereo~\cite{li2019stereo, chen2020dsgn, sun2020disp}, and multi-view ~\cite{zhang2024geobev, jiang2024fsd, zhang2023sa} approaches based on the number of onboard cameras. These methods leverage the rich color and texture information of image, along with its dense data representation, to achieve cost-effective 3D perception for autonomous driving. However, image-based methods are inherently limited by the lack of depth information, which restricts their performance. In contrast, point-cloud-based methods, which adopt voxel-based~\cite{zhou2023octr, yan2018second, deng2021voxel}, raw point cloud~\cite{shi2019pointrcnn, zhang2021pc, pan20213d}, or point-voxel hybrid~\cite{chen2019fast, shi2020points, shi2020pv} representations, can fully leverage the precise 3D geometric information provided by LiDAR, significantly improving perception capabilities. Considering that point-cloud-based methods suffer from sparse characteristic and lack rich semantic information, multimodal-fusion-based methods~\cite{liu2023bevfusion, zhang2022cat, yin2024fusion} further enhance performance by leveraging the complementary advantages of different modalities.

However, these single-agent-based 3D object detection methods are limited by sensor range and susceptible to occlusion, making them unable to detect objects that are further away or completely occluded. This paper focuses on collaborative-perception-based 3D object detection methods, which improve detection performance by supplementing the limitations of single-agent detection with information from other agents.

\subsection{Collaborative Perception}
Collaborative perception can be categorized into early collaboration, intermediate collaboration, and late collaboration based on the collaboration timing. Early collaboration~\cite{chen2019cooper, arnold2020cooperative, yu2022dair} shares raw perception data, providing good perception accuracy but with high bandwidth. Late collaboration~\cite{li2022v2x, xu2022opv2v} shares perception results, significantly reducing bandwidth, but leads to a decline in perception accuracy. Intermediate collaboration operates at the feature level and can achieve a better trade-off between accuracy and bandwidth by adjusting the intermediate features transmitted~\cite{guo2021coff, chen2019f, liu2020who2com, liu2020when2com, wang2020v2vnet, li2021learning, hu2022where2comm, xu2022v2x, xu2022opv2v, xu2023cobevt}, which is why it has been widely studied. Some of this research has focused on improving perception accuracy. FCooper~\cite{chen2019f} and CoFF~\cite{guo2021coff} used manual modeling to fuse multi-agent features. Who2com~\cite{liu2020who2com} and When2com~\cite{liu2020when2com} perform selective communication and use attention-based fusion. V2VNet~\cite{wang2020v2vnet} and DiscoNet~\cite{li2021learning} employ communication graph-based fusion methods. However, these methods typically transmit complete, uncompressed intermediate BEV features, leading to enormous bandwidth requirements, which makes them challenging to apply in practice.

To address the large bandwidth demand in collaborative perception, AttFuse~\cite{xu2022opv2v} was the first to use autoencoders to compress intermediate features along the channel dimension, which was later adopted by V2X-ViT~\cite{xu2022v2x}, CoBEVT~\cite{xu2023cobevt} and others. However, this method leads to significant accuracy degradation at high compression rates. Where2comm~\cite{hu2022where2comm} reduces bandwidth requirements by selecting sparse but important foreground regions for collaboration while maintaining perception accuracy. However, as bandwidth is further limited, perception accruacy still rapidly degrades, potentially even falling below the accuracy of simple late collaboration methods. To address this issue, this paper proposes a hybrid collaborative method based on both intermediate and late collaboration, which efficiently compensates for the intermediate collaborative results using late collaboration under bandwidth constraints. Furthermore, we also adopt and improve methods based on auto-encoders and information selection to save bandwidth while maintaining accuracy.

\section{Method}
\label{sec:method}

\begin{figure*}[t]
  \centering
  \includegraphics[width=0.98\textwidth]{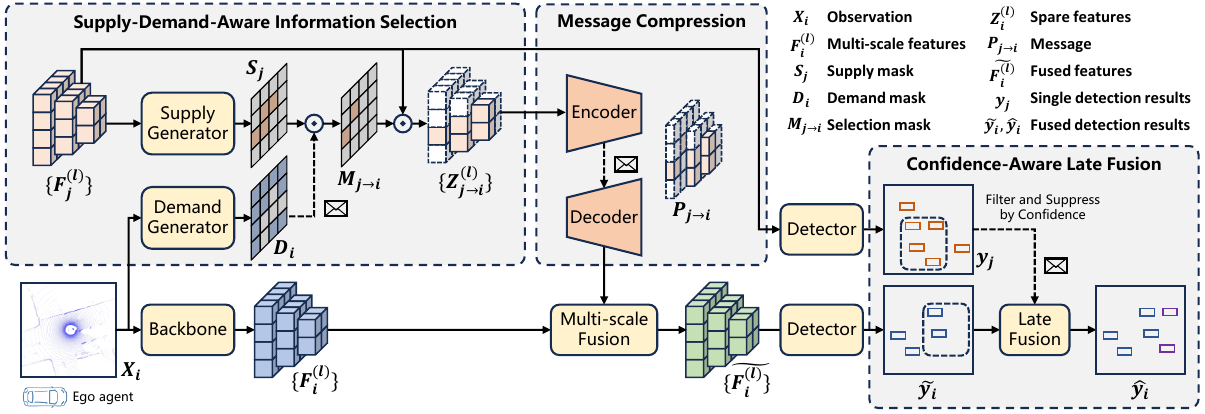}
  \caption{The overall architecture of \mymethodname. The Supply-Demand-Aware Information Selection module selects sparse but important information, which is then further compressed by the Message Compression module to achieve efficient communication. Confidence-Aware Late Fusion compensates for the intermediate fusion detection results to improve accuracy.}
  \label{fig:framework}
  \vspace{-8pt}
\end{figure*}

\subsection{Problem Definition}
In this paper, we consider the problem of collaborative perception with $N$ agents. Let $X_i$ and $y_i$ represent the raw observation and the corresponding ground truth supervision of the $i$-th agent, respectively, and let $P_{j\rightarrow i}$ be the message sent from agent $j$ to agent $i$. In the collaborative perception, agent $i$ aggregates its own observations and the messages $\{P_{j\rightarrow i}\}_{j=1}^{N}$ sent from other agents to perform 3D object detection task. Our goal is to maximize the collaborative perception 3D object detection accuracy while ensuring that each agent has a communication budget $B$:
\begin{align}
\xi_{\Phi}(B) = \underset{\theta}{\text{arg max}} &\sum_{i=1}^{N}{g(\Phi_{\theta}(X_i,\{P_{j\rightarrow i}\}_{j=1}^{N}),y_i)},\\
\text{s.t.} &\sum_{j=1}^{N}{|P_{j\rightarrow i}|} \le B
\end{align}
where $\Phi_{\theta}$ represents the collaborative perception 3D object detection model, $\theta$ denotes the model parameters, $|P_{j\rightarrow i}|$ represents the communication volume of the message sent from agent $j$ to agent $i$, and $g(\cdot , \cdot)$ denotes the 3D object detection evaluation metric.

\subsection{Overall Architecture}
The overall architecture of the proposed \mymethodname~is shown in Fig.~\ref{fig:framework}. Each agent first processes its locally observed point cloud $X_i$ through a backbone network based on PointPillar~\cite{lang2019pointpillars} and a demand generator to obtain multi-scale BEV features $\{F_{i}^{(l)}\}_{l=1,2,...,L}$ and a demand mask $D_i$, respectively. Considering the collaboration between Ego agent $i$ and collaborating agent $j$, agent $j$ generates a supply mask $S_j$ from its multi-scale features $\{F_{j}^{(l)}\}_{l=1,2,..,L}$ via a supply generator, and multiplies it element-wise with the received demand matrix to obtain the supply-demand mask $M_{j\rightarrow i}$. Agent $j$ then performs supply-demand-aware information selection by multiplying $\{F_{j}^{(l)}\}_{l=1,2,..,L}$ with $M_{j\rightarrow i}$ element-wise to obtain sparse spatial features $\{Z_{j}^{(l)}\}_{l=1,2,..,L}$. Subsequently, agent $j$ compresses the features through an autoencoder, sending the non-zero parts of the features along with their corresponding coordinates as the message $P_{j\rightarrow i}$ to agent $i$. 

Upon receiving $P_{j\rightarrow i}$, agent $i$ first decodes the features to restore their dimensions and then fuses them with its local features $\{F_{i}^{(l)}\}_{l=1,2,...,L}$ across multiple scales to obtain the fused features $\{\tilde{F_{i}^{(l)}}\}_{l=1,2,...,L}$, which are then passed to the detection head for intermediate collaborative detection results $\tilde{y_i}$. Afterward, we apply confidence-aware late fusion: agent $j$ filters and suppresses its own detection results $y_j$ based on confidence and sends them to agent $i$ for late fusion, yielding the final hybrid collaborative perception detection results $\hat{y_i}$.

\subsection{Supply-Demand-Aware Information Selection}
Previous methods such as Where2comm~\cite{hu2022where2comm} and How2comm~\cite{yang2024how2comm} generally use symmetric supply-demand relationships to select sparse features for collaboration, believing that areas with high foreground confidence in the collaborating agent’s view should be provided, and conversely, areas with low foreground confidence in the Ego agent’s view need collaboration. However, we believe that areas with low foreground confidence from the Ego's perspective can be divided into areas that are hard to observe and those that can be observed but belong to the background. The latter do not require collaboration. Based on this insight, we propose a novel supply-demand-aware information selection method.

The demand mask $D_i$ indicates where agent $i$ needs information from collaborating agents. Intuitively, the agent requires information from areas that are distant or occluded, which have the common characteristic of having low point cloud density or no point cloud at all. For agent $i$, we consider using the number of point clouds in each pillar to represent point cloud density, and we map it to the range $[0,1]$, i.e., $A_{i} \in [0,1]^{H\times W}$, where $H$ and $W$ represent the number of Pillars along the height and width dimensions. We then select areas where the point cloud density is below a threshold $\epsilon_a$ to obtain the demand mask for agent $i$, $D_{i}=A_{i}< \epsilon_{a}\in \{0,1\}^{H\times W}$. The demand mask indicates where agent $i$ has poor perception and needs collaborative information from other agents. Filtering information from other agents using the demand mask not only helps save bandwidth but also avoids interference from other agents' information in well-perceived areas.

For the object detection task, foreground information is more valuable. Providing sparse foreground features can effectively assist other agents in supplementing undetected and incomplete targets while using less bandwidth. Following previous work~\cite{hu2022where2comm}, we use the spatial confidence map $C_{i}\in [0,1]^{H\times W}$ output by the detection head to select potential foreground areas. We use a 
supply threshold $\epsilon_{c}$ to obtain the supply mask $S_{i}^{(l)}=C_{i}>\epsilon_{c}\in \{0,1\}^{H\times W}$. By adjusting the threshold, we can dynamically adjust the bandwidth used for collaborative perception to adapt to varying communication conditions.

During collaboration, agent $j$ generates a binary supply-demand selection mask $M_{j\rightarrow i} = D_i \odot S_j \in \{0, 1\}^{H\times W}$ based on its supply mask $S_j$ and agent $i$'s demand mask $D_i$, and samples it to multiply element-wise with multi-scale BEV features $\{F_j^{(l)}\}_{l=1,2,...,L}$, obtaining sparse features $\{Z_{j\rightarrow i}^{(l)}\}_{l=1,2,...,L}$. During communication, only the non-zero parts and their corresponding coordinates need to be transmitted.

\begin{figure}
  \centering
  \includegraphics[width=0.49\textwidth]{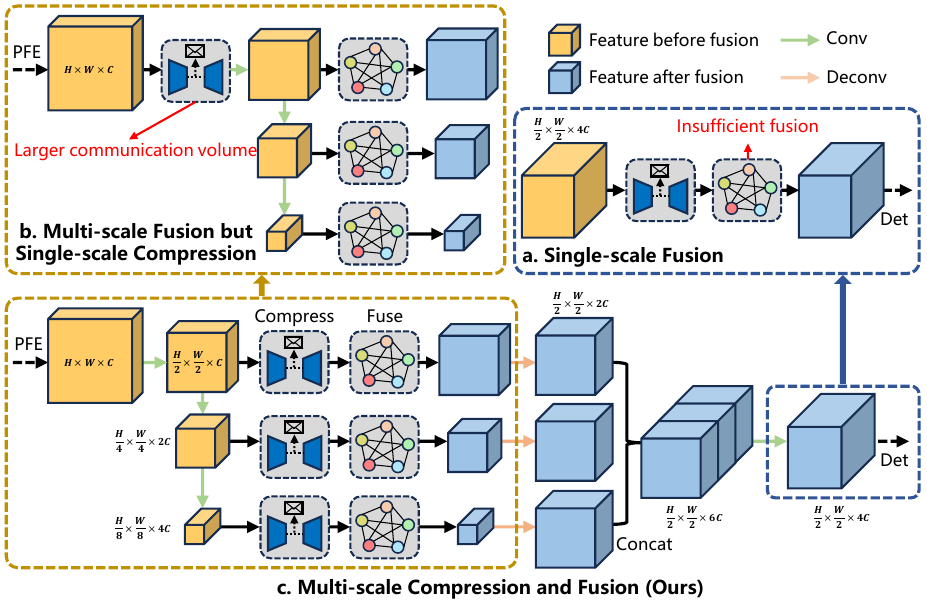}
  \caption{Comparison of our multi-scale compression and fusion with other methods. It can achieve thorough fusion with smaller communication volume.}
  \label{fig:compress}
  \vspace{-8pt}
\end{figure}

\subsection{Message Compression and Fusion}
To further reduce communication bandwidth, we also use autoencoders to compress intermediate features along the channel dimension before communication, while making further improvements to existing compression and fusion schemes. As shown in Fig.~\ref{fig:compress}, traditional single-scale fusion schemes~\cite{xu2022opv2v, xu2022v2x} only compress and fuse a single-layer BEV feature map before passing it to the detection head, resulting in insufficient fusion. CoAlign~\cite{lu2023robust} proposed a solution that performs layer-wise fusion on multi-scale features, addressing the issue of insufficient fusion. However, it compresses large-scale single-layer features extracted after Pillar Feature Extraction (PFE), which leads to higher bandwidth. Therefore, we propose a multi-scale compression and fusion approach, where separate autoencoders are designed for each scale of features to perform compression, followed by layer-wise fusion. This approach enables more thorough fusion while using less bandwidth.

For message fusion, we use Max fusion, which has two main advantages: i) it is computationally simple and efficient, with the complexity increasing linearly with the number of agents; ii) max fusion selects the maximum value of features from multiple agents, achieving information complementarity. Considering the collaboration between Ego agent $i$ and collaborating agent $j$, the process of message compression and fusion can be expressed as:
\begin{align}
Z_{j\rightarrow i}^{'(l)} &= f_{encode}^{(l)}(Z_{j\rightarrow i}^{(l)}) \in \mathbb{R}^{\frac{C_{l}}{c_0} \times H_{l} \times W_{l}} \\
F_{j\rightarrow i}^{(l)} &= f_{decode}^{(l)}(Z_{j\rightarrow i}^{'(l)}) \in \mathbb{R}^{C_{l} \times H_{l} \times W_{l}} \\
F_{j\rightarrow i}^{'(l)} &= f_{transform}(F_{j\rightarrow i}^{(l)}, \xi_{j\rightarrow i}) \in \mathbb{R}^{C_{l} \times H_{l} \times W_{l}} \\
\tilde{F_{i}^{(l)}} &= \max{(F_{i}^{(l)}, \{F_{j\rightarrow i}^{'(l)}\}_{j\ne i})} \in \mathbb{R}^{C_{l} \times H_{l} \times W_{l}}
\end{align}
where $Z_{j\rightarrow i}^{(l)} \in \mathbb{R}^{C_{l} \times H_{l} \times W_{l}}$ represents the sparse features selected based on supply-demand relationships from the previous stage, and $C_l, H_l, W_l$ denote the channel, height, and width dimensions of the $l$-th layer feature map. $f_{encode}^{(l)}, f_{decode}^{(l)}$ represent the encoder and decoder of the $l$-th layer autoencoder, $c_0$ is the compression ratio in the channel dimension, and $f_{transform}$ represents coordinate transformation. During communication, $\{Z_{j\rightarrow i}^{'(l)}\}_{l=1,2,..,L}$ is first converted from float32 to float16, and then only the non-zero parts and corresponding coordinates are transmitted to save bandwidth. Upon receiving the message, agent $i$ decodes the feature dimensions, then aligns the features to its own coordinate system using the coordinate transformation matrix $\xi_{j\rightarrow i}$ to obtain $F_{j\rightarrow i}^{'(l)}$, and finally fuses its own features with the collaborating agent’s features using max fusion to obtain the fused features $\tilde{F_{i}^{(l)}}$.

\subsection{Confidence-Aware Late Fusion}
Existing collaborative perception methods focus on the singular intermediate collaboration architecture and neglect the advantages of late collaboration. As shown in Table~\ref{tab:result1}, the experimental results indicate that late fusion can demonstrate acceptable perception accuracy with extremely low bandwidth on the OPV2V~\cite{xu2022opv2v} and V2XSim~\cite{li2022v2x} datasets. Therefore, late fusion can be used to compensate for the results of intermediate fusion, achieving a better accuracy-bandwidth trade-off.

Naive late collaboration methods~\cite{li2022v2x, xu2022opv2v} directly merge results from other agents and then apply NMS (Non-Maximum Suppression) to deduplicate and obtain final results. This approach can effectively improve recall for collaborative perception object detection; however, during the merging process, it may introduce some low-confidence false positives, and suboptimal detection results from collaborating agents may override the Ego agent’s detection results, lowering precision and consequently decreasing final AP. The lower AP in the late fusion method in Table~\ref{tab:result1} on the DAIR-V2X~\cite{yu2022dair} dataset is due to this issue.

\begin{figure}
  \centering
  \includegraphics[width=0.49\textwidth]{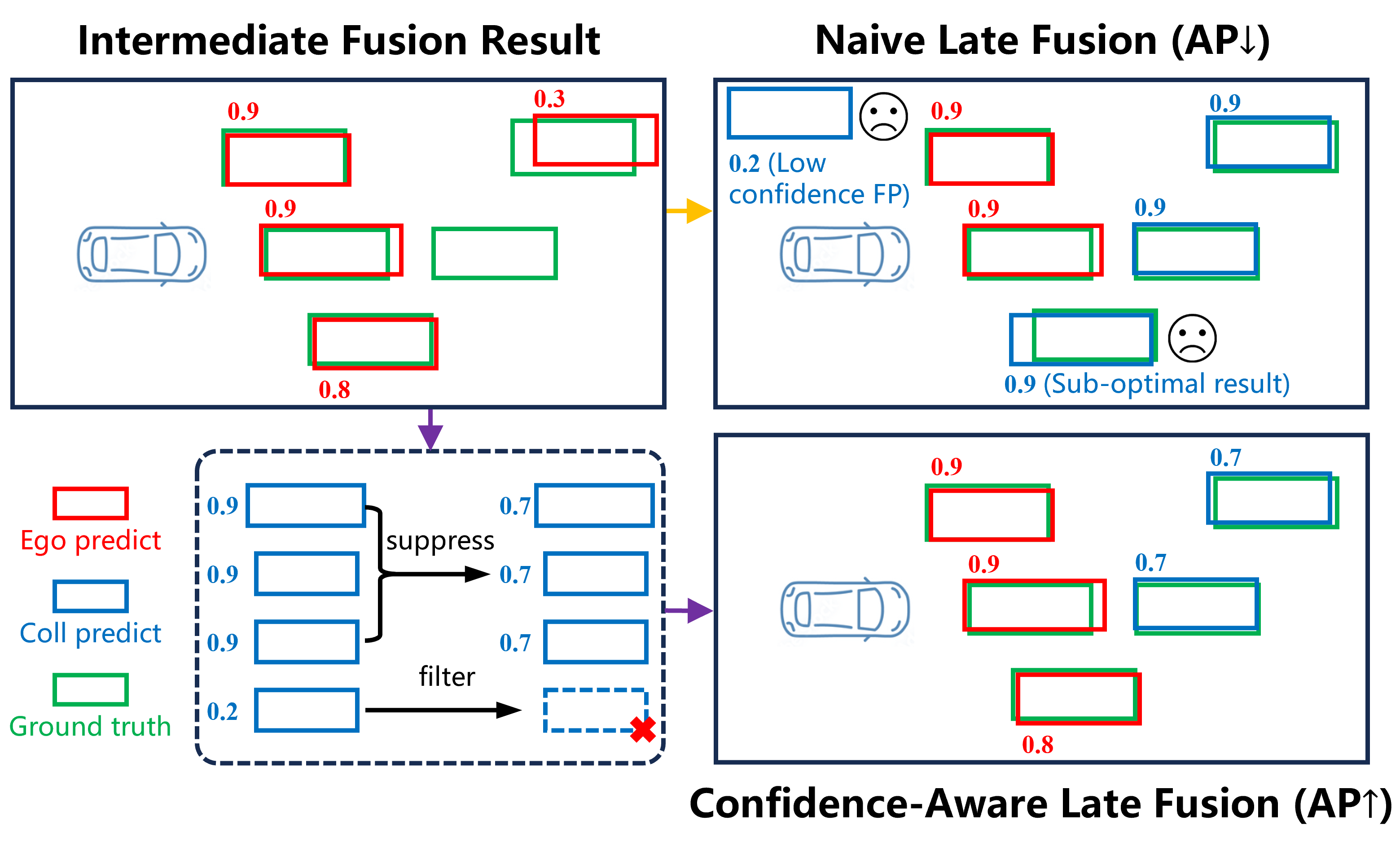}
  \caption{Our confidence-aware late fusion. It filters detection results based on confidence and suppresses suboptimal results from collaborative agents, improving overall detection accuracy.}
  \label{fig:late-fusion}
  \vspace{-9pt}
\end{figure}

To address this problem, we propose a new confidence-aware late fusion method. As shown in Fig.~\ref{fig:late-fusion}, during late fusion, we first filter based on the confidence of the target boxes, discarding target boxes from other agents with confidence lower than $\epsilon_{l}$. Furthermore, considering that if both the Ego agent and other agents detect the same target, the detection result from the Ego agent, which has undergone intermediate fusion, is of higher quality. Therefore, we suppress the detection results from other agents to prevent their lower-quality results from degrading the Ego agent's better detection results. Specifically, before merging the detection results, we multiply the confidence of target boxes from other agents by a coefficient $\beta \in (0, 1)$.

\begin{table*}[t]
\centering
\resizebox{\textwidth}{!}{
\begin{tabular}{c|c|ccc|ccc|ccc}
\hline
\multirow{2}{*}{Setting}         & \multirow{2}{*}{Method} 
                                & \multicolumn{3}{c|}{OPV2V~\cite{xu2022opv2v}}
                                & \multicolumn{3}{c|}{V2XSim~\cite{li2022v2x}}
                                & \multicolumn{3}{c}{DAIR-V2X~\cite{yu2022dair}} \\ \cline{3-11} 
                                 &                         & AP@0.5$\uparrow$  & AP@0.7$\uparrow$  & BD$\downarrow$
                                                           & AP@0.5$\uparrow$  & AP@0.7$\uparrow$  & BD$\downarrow$
                                                           & AP@0.5$\uparrow$  & AP@0.7$\uparrow$  & BD$\downarrow$   \\ \hline
\multirow{3}{*}{Basic}
& No Fusion                         &79.78\%&67.16\%& 0.0 Mbps   &70.31\%&58.55\%&    0.0 Mbps&66.51\%&55.46\%&    0.0 Mbps  \\
& Early Fusion                      &95.05\%&88.98\%& 83.1 Mbps  &95.68\%&88.03\%&   55.5 Mbps&74.52\%&59.22\%&   50.2 Mbps  \\
& Late Fusion                       &94.70\%&87.18\%& 0.2 Mbps   &86.88\%&78.13\%&    0.1 Mbps&67.86\%&50.47\%&    0.2 Mbps  \\ \hline
\multirow{9}{*}{No Limit}
& When2com~\cite{liu2020when2com}   &91.75\%&81.77\%&1,320.0 Mbps&72.65\%&62.92\%&  250.0 Mbps&64.08\%&49.14\%&  984.4 Mbps  \\
& FCooper~\cite{chen2019f}          &90.06\%&74.03\%&2,640.0 Mbps&72.96\%&57.61\%&  500.0 Mbps&74.58\%&56.47\%&1,968.8 Mbps  \\
& AttFuse~\cite{xu2022opv2v}        &94.31\%&82.03\%&2,640.0 Mbps&78.06\%&64.84\%&  500.0 Mbps&73.80\%&56.86\%&1,968.8 Mbps  \\
& V2VNet~\cite{wang2020v2vnet}      &96.66\%&92.44\%&5,280.0 Mbps&88.97\%&85.18\%&1,000.0 Mbps&66.63\%&47.39\%&3,937.5 Mbps  \\
& DiscoNet~\cite{li2021learning}    &90.93\%&78.90\%&2,640.0 Mbps&77.34\%&68.77\%&  500.0 Mbps&73.58\%&58.45\%&1,968.8 Mbps  \\
& V2XViT~\cite{xu2022v2x}           &95.87\%&89.88\%&2,640.0 Mbps&89.01\%&80.26\%&  500.0 Mbps&76.68\%&57.57\%&1,920.0 Mbps  \\
& Where2comm~\cite{hu2022where2comm}&95.59\%&91.39\%&   48.7 Mbps&88.18\%&83.66\%&   27.5 Mbps&76.13\%&60.16\%&  172.3 Mbps  \\
& CoAlign~\cite{lu2023robust}       &96.63\%&92.63\%&2,640.0 Mbps&88.87\%&85.23\%&  500.0 Mbps&\textbf{78.06\%}&63.09\%&1,968.8 Mbps  \\
& CoSDH                              &\textbf{96.83\%}&\textbf{92.99\%}&  \textbf{13.4 Mbps} &\textbf{89.23\%}&\textbf{86.31\%}&   \textbf{1.1 Mbps}&76.75\%&\textbf{63.85\%}&   \textbf{7.1 Mbps}  \\ \hline
\multirow{9}{*}{BD $\le$ 6.75 Mbps}
& When2com~\cite{liu2020when2com}   &79.97\%&50.88\%& 5.2 Mbps   &62.02\%&43.37\%&    4.0 Mbps&62.37\%&44.01\%&    3.8 Mbps  \\
& FCooper~\cite{chen2019f}          &90.37\%&72.92\%& 5.2 Mbps   &76.52\%&61.83\%&    4.0 Mbps&67.71\%&47.65\%&    3.8 Mbps  \\
& AttFuse~\cite{xu2022opv2v}        &93.45\%&80.02\%& 5.2 Mbps   &84.58\%&71.36\%&    4.0 Mbps&71.06\%&49.57\%&    3.8 Mbps  \\
& V2VNet~\cite{wang2020v2vnet}      &95.71\%&87.16\%& 5.2 Mbps   &85.66\%&73.72\%&    4.0 Mbps&66.49\%&45.61\%&    3.8 Mbps  \\
& DiscoNet~\cite{li2021learning}    &90.00\%&76.72\%& 5.2 Mbps   &78.04\%&68.18\%&    4.0 Mbps&70.83\%&53.40\%&    3.8 Mbps  \\
& V2XViT~\cite{xu2022v2x}           &95.85\%&87.13\%& 5.2 Mbps   &88.94\%&81.47\%&    4.0 Mbps&71.00\%&52.78\%&    3.8 Mbps  \\
& Where2comm~\cite{hu2022where2comm}&94.91\%&89.86\%& 5.4 Mbps   &87.51\%&82.12\%&    4.7 Mbps&74.98\%&59.51\%&    5.3 Mbps  \\
& CoAlign~\cite{lu2023robust}       &94.10\%&85.99\%& 5.2 Mbps   &88.01\%&83.97\%&    4.0 Mbps&75.26\%&60.19\%&    3.8 Mbps  \\
& CoSDH                              &\textbf{96.75\%}&\textbf{92.92\%}& \textbf{2.0 Mbps}   &\textbf{89.23\%}&\textbf{86.31\%}&    \textbf{1.1 Mbps}&\textbf{76.47\%}&\textbf{63.76\%}&    \textbf{1.4 Mbps}  \\ \hline
\end{tabular}
}
\caption{Comparison of detection accuracy and bandwidth of different methods on OPV2V~\cite{xu2022opv2v}, V2XSim~\cite{li2022v2x}, and DAIR-V2X~\cite{yu2022dair} datasets. ``BD'' represents the bandwidth required for each collaborative agent, assuming Ego agent collaborates with up to 4 agents and the detection frequency is 10Hz. ``BD$\le$6.75 Mbps'' is used to simulate real-world communication limits, assuming a total communication bandwidth of 27 Mbps, with each collaborating agent's bandwidth consumption limited to less than 6.75 Mbps. For intermediate collaboration methods without information selection, we provide a compressed version using autoencoders to meet the bandwidth constraints. For intermediate collaboration methods with information selection, the selection ratio is adjusted to meet the bandwidth constraints.}
\label{tab:result1}    
\vspace{-8pt}
\end{table*}
\section{Experiments}
\label{sec:experiments}

\subsection{Datasets and Experimental Settings}

\textbf{Datasets.} We evaluate the proposed CoSDH against other methods on three different collaborative perception datasets (\textbf{OPV2V}~\cite{xu2022opv2v}, \textbf{V2XSim}~\cite{li2022v2x}, and \textbf{DAIR-V2X}~\cite{yu2022dair}) for LiDAR-based 3D object detection. The datasets include both simulated and real-world scenarios, and cover two types of collaboration: V2V (Vehicle to Vehicle) and V2I (Vehicle to Infrastructure).

\textbf{Evaluation Metrics}
\label{sec:metrics}
We use the average precision (AP) with intersection-over-union (IoU) thresholds of 0.5 and 0.7 to evaluate the performance of different methods on 3D object detection. We assume the target detection frequency is 10Hz and calculate the communication bandwidth based on the average data transmitted by each collaborative agent to the Ego agent, in order to evaluate the communication cost of different methods. Specifically, we consider the bandwidth limitations in real-world collaborative perception scenarios, setting the vehicle's communication rate to $27$ Mbps~\cite{arena2019overview, lozano2019review}. Considering the typical case where the Ego agent collaborates with up to 4 other agents~\cite{xu2022opv2v}, the bandwidth limit for each collaborative agent is $27/4=6.75$ Mbps.

\textbf{Implementation} Our experiments are based on the OpenCOOD~\cite{xu2022opv2v} framework, using PointPillar~\cite{lang2019pointpillars} as the encoder with a grid size of $(0.4m, 0.4m)$, and a maximum of 32 points per Pillar. For our method, we set the number of intermediate feature layers to $L=3$, the demand threshold $\epsilon_{a}=4/32=0.125$, and the supply threshold $\epsilon_{c}=0.01$. For the OPV2V~\cite{xu2022opv2v} and DAIR-V2X~\cite{yu2022dair} datasets, the compression rate is $c_0=16$, and for the V2XSim~\cite{li2022v2x} dataset, the compression rate is $c_0=8$ due to its smaller perception range and lower inherent bandwidth requirement. The late fusion threshold is set to $\epsilon_{l}=0.3$, with a suppression coefficient $\beta=0.9$ for OPV2V and V2XSim dataset and $\beta=0.8$ for DAIR-V2X dataset because of its difficulty. In late fusion, the dense prediction results before NMS are transmitted, as this reduces the computational burden on the collaborating agents. The experiments use the Adam~\cite{kingma2014adam} optimizer, with an initial learning rate set between 0.0001 and 0.002 based on the model's testing complexity to ensure proper training. The maximum number of collaborating agents is set to 5. The number of training epochs is set to 40 to ensure model convergence. Other experimental parameters are kept consistent with the OpenCOOD framework. All methods are trained on four NVIDIA GeForce RTX 3090 GPUs.

\begin{figure*}[t]
  \centering
  \includegraphics[width=\textwidth]{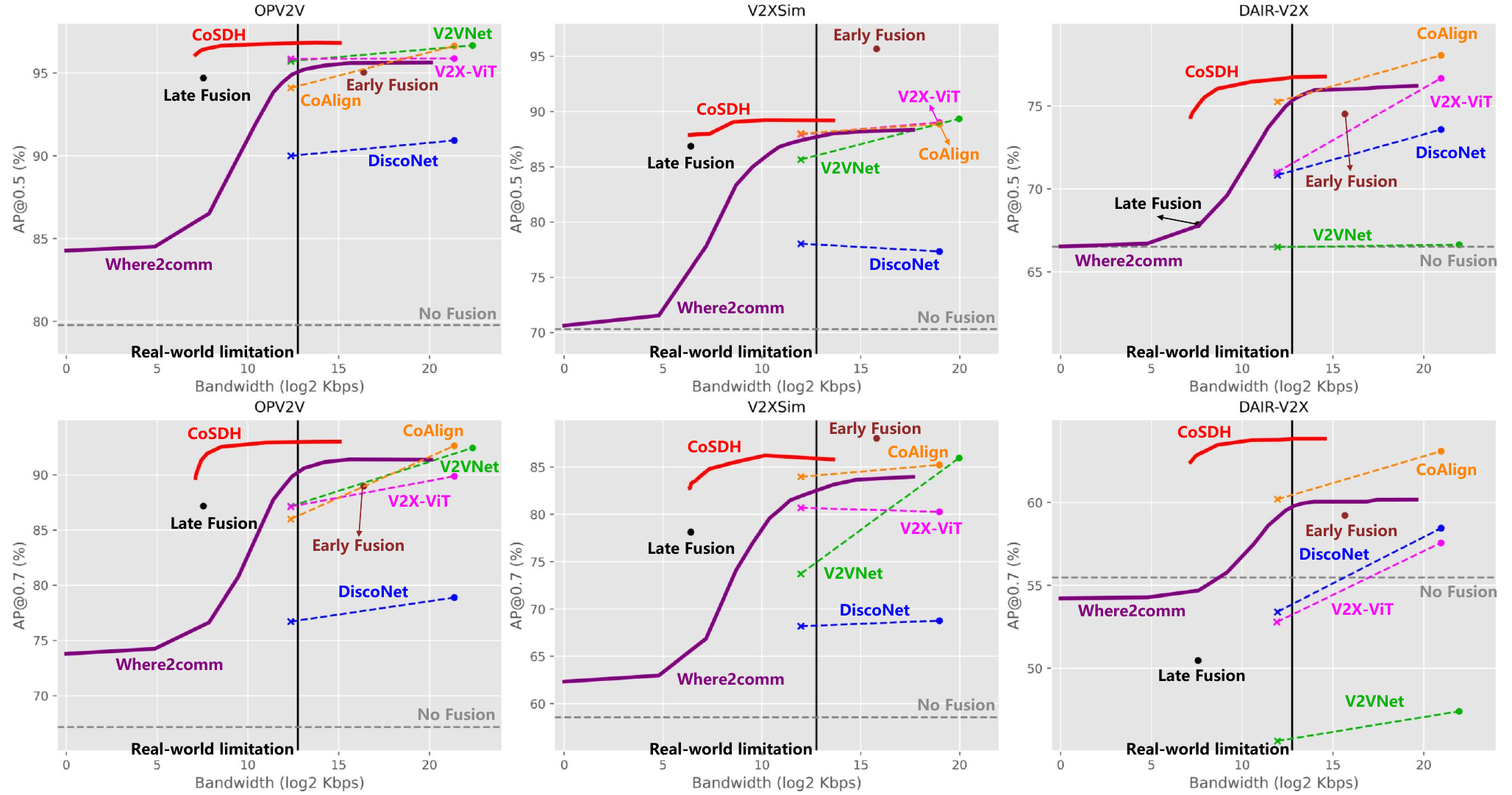}
  \caption{Comparison of the trade-off between detection accuracy and bandwidth of different methods on OPV2V~\cite{xu2022opv2v}, V2XSim~\cite{li2022v2x} and DAIR-V2X~\cite{yu2022dair} datasets, \mymethodname~achieves the best accuracy-bandwidth trade-off. The real-world limitation refers to the total bandwidth limit of 27 Mbps, which means that each collaborative agent does not exceed 6.75 Mbps.}
  \label{fig:trade-off}
  \vspace{-8pt}
\end{figure*}

\subsection{Quantitative Evaluation}
\textbf{Benchmark Comparison}. Table~\ref{tab:result1} presents the collaborative 3D object detection accuracy and required bandwidth of the proposed \mymethodname~ compared to previous methods across different datasets. Experimental results show that, with the default uncompressed settings, \mymethodname~ achieves the highest accuracy on the OPV2V~\cite{xu2022opv2v} and V2XSim~\cite{li2022v2x} datasets while requiring only about 1/100 to 1/1000 of the bandwidth compared to other non-communication-efficient methods. Although the AP@0.5 of \mymethodname~ on the DAIR-V2X~\cite{yu2022dair} dataset is slightly lower than that of CoAlign~\cite{lu2023robust}, it achieves a higher AP@0.7 with only about 1/300 of the bandwidth. Both our ~\mymethodname~ and Where2comm~\cite{hu2022where2comm} are communication-efficient methods based on information selection, which enable dynamic accuracy-bandwidth trade-offs by adjusting the selection ratio. The table shows accuracy at a specific bandwidth, and a more detailed comparison of the accuracy-bandwidth curves is provided in the ``Accuracy-Bandwidth Trade-Off Comparison'' part.

Furthermore, we simulate real-world communication rate limitations. Result shows that \mymethodname~ achieves improvements in AP@0.7 of 3.06\%/2.34\%/3.75\% on OPV2V/V2XSim/DAIR-V2X compared to the previous best methods, while using less bandwidth. When comparing the scenarios with and without bandwidth limitations on the OPV2V and DAIR-V2X datasets, we found that ~\mymethodname~ achieves less than a 0.3\% decrease in AP under conditions where bandwidth is reduced by 80\% to 85\%, exhibiting less accuracy degradation than Where2comm. Most other methods also show varying degrees of accuracy degradation under bandwidth constraints. Interestingly, under bandwidth limitations, some methods show an improvement in accuracy on certain datasets after using autoencoders to compress intermediate features. For example, FCooper~\cite{chen2019f} shows improved accuracy on the V2XSim dataset, which may be due to the model's initially poor performance. The compressed version, with the added autoencoder, increases the model's parameter count, thereby enriching its expressive capability.

\textbf{Accuracy-Bandwidth Trade-Off Comparison}. Fig.~\ref{fig:trade-off} shows the accuracy-bandwidth trade-off of the proposed \mymethodname~ and previous advanced methods across different datasets. When the bandwidth grater than 1 Mbps, Where2comm~\cite{hu2022where2comm} demonstrates a good enough accuracy-bandwidth trade-off, maintaining high detection accuracy as bandwidth decreases. However, as bandwidth further decreases, its accuracy quickly declines, even falling below that of late fusion method. \mymethodname~ can leverage late fusion at low bandwidths to maintain high accuracy, achieving a better performance-bandwidth trade-off. Notably, on the DAIR-V2X dataset, the late fusion method performs worse than the no-fusion case because the naive late fusion method unselectively merges detection results from other agents, introducing a large number of low-quality detections. \mymethodname~ uses a confidence-aware late fusion method to select higher-quality detection results, improving this situation and enhancing perception accuracy. We also observe that the AP@0.7 of \mymethodname~ on the V2XSim dataset initially increases slightly as bandwidth decreases before subsequently declining, indicating that selecting key collaboration areas can help reduce interference from other regions to some extent, thereby slightly improving detection accuracy.

\subsection{Qualitative Evaluation}

\begin{figure}
  \centering
  \includegraphics[width=0.495\textwidth]{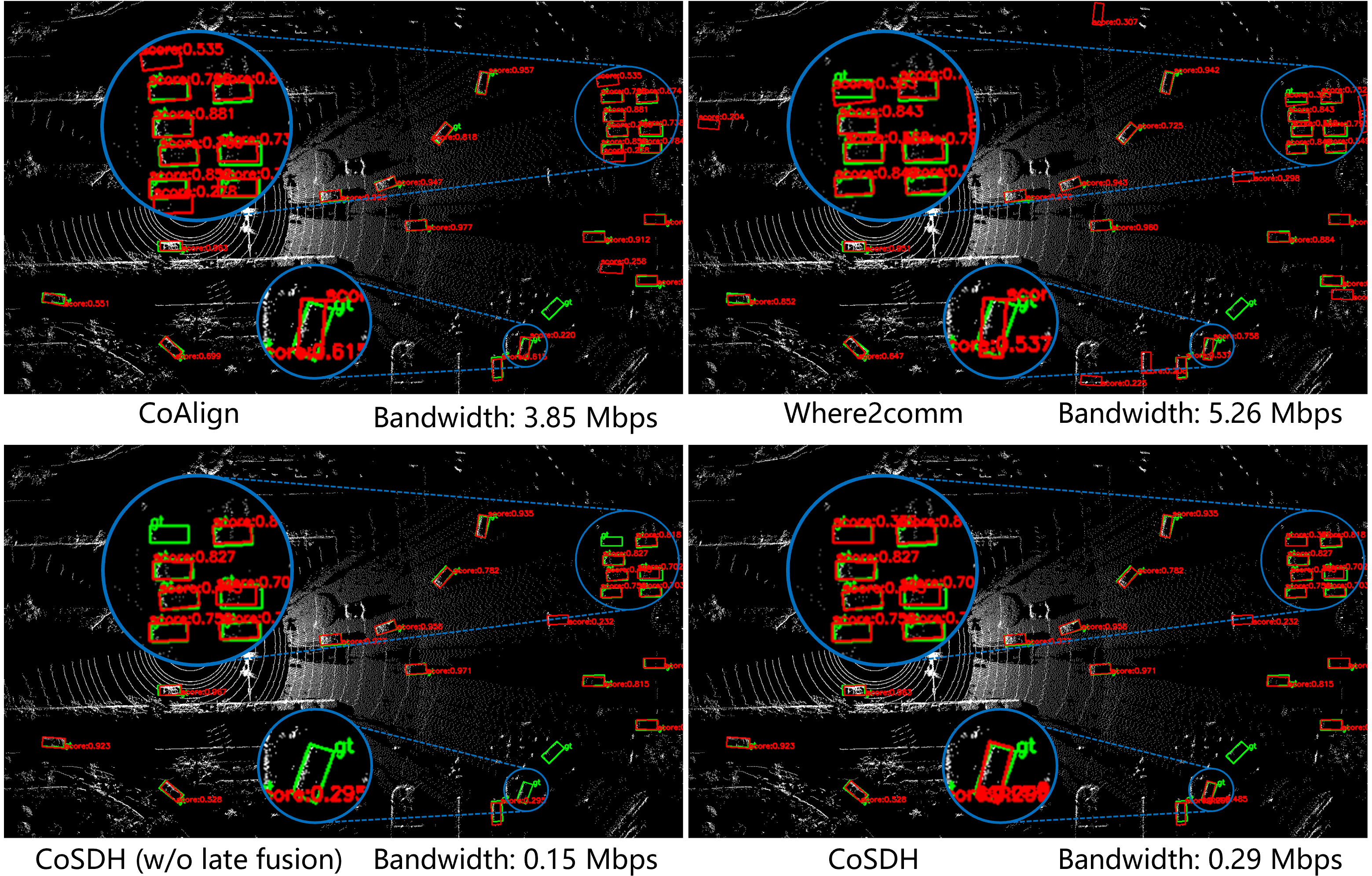}
  \caption{Visualization of detection results under real-world communication rate limitations on the DAIR-V2X~\cite{yu2022dair} dataset. {\color{green} Green} represents ground truth box, and {\color{red} red} represents predicted box.}
  \label{fig:vis1}
  \vspace{-8pt}
\end{figure}

Fig.~\ref{fig:vis1} shows the visualization of detection results for our method compared to other methods under simulated real-world communication rate limitations on the DAIR-V2X dataset. A comparison of the detection results reveals that, while using less bandwidth, our method achieves the same recall rate as CoAlign~\cite{lu2023robust} and Where2comm~\cite{hu2022where2comm}, with fewer false positive predictions and more accurate object localization, demonstrating that our method performs better under low-bandwidth conditions. Comparing the two images at bottom, it can be seen that under low bandwidth conditions, late fusion effectively compensates for the results of intermediate fusion, improving the recall rate of objects.

\subsection{Ablation Studies}
\label{sec:ablation}

Table~\ref{tab:ablation1} presents the results of the ablation study on various modules of the proposed method using the OPV2V~\cite{xu2022opv2v} dataset. The results show that after using the autoencoder for compression, the bandwidth is reduced by a factor of $c_0$, and there is no significant loss in perception accuracy, even with a slight improvement in AP@0.7. Converting intermediate features to float16 can nearly halve the bandwidth without significant loss, demonstrating the effectiveness of our proposed message compression module. After performing information selection based on the supply mask, perception accuracy slightly decreases, but the bandwidth is reduced by about 95\%, which is a worthwhile trade-off. Further using the demand mask results in no significant decrease in accuracy but reduces bandwidth by about 10\%. This demonstrates that our supply-demand-aware information selection can reduce bandwidth while maintaining accuracy.

After adding late fusion, accuracy improves at a relatively small bandwidth cost. However, since this phase uses more bandwidth, the improvement in accuracy is not significant. Table~\ref{tab:ablation2} shows the accuracy improvement due to late collaboration under different bandwidth conditions. It can be observed that late collaboration provides more accuracy improvements under lower bandwidth conditions. For example, when the bandwidth used for intermediate fusion is 0.05 Mbps, late collaboration with a bandwidth cost of 0.14 Mbps leads to a 3\% improvement in AP@0.5 and a 4\% improvement in AP@0.7. This is one of the key reasons why our method achieves significantly higher detection accuracy under lower bandwidth compared to Where2comm~\cite{hu2022where2comm}.

\begin{table}[]
\centering
\resizebox{0.49\textwidth}{!}{
\begin{tabular}{cc|cc|c|ccc}
\hline
\multicolumn{2}{c|}{Compression} & \multicolumn{2}{c|}{Selection} & \multirow{2}{*}{Late Fusion} & \multirow{2}{*}{AP@0.5$\uparrow$} & \multirow{2}{*}{AP@0.7$\uparrow$} & \multirow{2}{*}{BD$\downarrow$} \\ \cline{1-4}
Autoencoder        & FP16        & Supply               & Demand              &                              &                         &                         &                     \\ \hline

    &   &   &   &   &96.62\%&92.62\%&1,155.00 Mbps \\
$\checkmark$&   &   &   &   &96.59\%&92.99\%&72.19 Mbps \\
$\checkmark$&$\checkmark$&   &   &   &96.59\%&92.99\%&36.09 Mbps \\
$\checkmark$&$\checkmark$&$\checkmark$&   &   &96.30\%&92.62\%&1.99 Mbps \\
$\checkmark$&$\checkmark$&$\checkmark$&$\checkmark$&   &96.31\%&92.60\%&1.82 Mbps \\
$\checkmark$&$\checkmark$&$\checkmark$&$\checkmark$&$\checkmark$&96.75\%&92.92\%&1.97 Mbps \\ \hline
\end{tabular}
}
\caption{Ablation study of the modules in \mymethodname~ on the OPV2V dataset. ``Autoencoder'' refers to the use of autoencoders to compress features along the channel dimension, with a compression ratio of $c_0=16$. ``FP16'' refers to converting features from float32 to float16 for transmission. ``Supply'' and ``Demand'' refer to using supply and demand masks for information selection, and ``Late Fusion'' refers to applying confidence-aware late fusion.}
\label{tab:ablation1}
\end{table}

\begin{table}[]
\centering
\resizebox{0.49\textwidth}{!}{
\begin{tabular}{c|c|ccccc}
\hline
Late Fusion              & $\epsilon_c$     & 0.01       & 0.02      & 0.03      & 0.05      & 0.07      \\ \hline
\multirow{3}{*}{} & AP@0.5$\uparrow$ & 96.59\%    & 96.31\%   & 96.19\%   & 95.18\%   & 93.35\%   \\
                         & AP@0.7$\uparrow$ & 92.95\%    & 92.60\%   & 92.27\%   & 90.67\%   & 87.60\%   \\
                         & BD$\downarrow$   & 13.26 Mbps & 1.82 Mbps & 0.54 Mbps & 0.12 Mbps & 0.05 Mbps \\ \hline
\multirow{3}{*}{$\checkmark$}    & AP@0.5$\uparrow$ & 96.83\%    & 96.75\%   & 96.72\%   & 96.54\%   & 96.41\%   \\
                         & AP@0.7$\uparrow$ & 92.99\%    & 92.92\%   & 92.73\%   & 92.23\%   & 91.68\%   \\
                         & BD$\downarrow$   & 13.40 Mbps & 1.97 Mbps & 0.68 Mbps & 0.26 Mbps & 0.19 Mbps \\ \hline
\end{tabular}
}
\caption{Ablation study of confidence-aware late fusion under different bandwidths on the OPV2V~\cite{xu2022opv2v} dataset. The table shows the impact of late fusion on accuracy and bandwidth with different values of $\epsilon_c$.}
\label{tab:ablation2}
\vspace{-8pt}
\end{table}

\section{Conclusion}
\label{sec:conclusion}

In this paper, we propose \mymethodname, a novel communication-efficient collaborative perception framework for 3D object detection. By finely modeling the supply-demand relationship between agents, it selects key and sparse regions for collaboration. Additionally, we innovatively incorporate confidence-aware late fusion on top of intermediate collaboration to form an intermediate-late hybrid collaborative perception method. Experiments on multiple datasets show that our method offers a better accuracy-bandwidth trade-off, demonstrating outstanding accuracy under bandwidth constraints close to real-world communication limits, making it highly valuable for practical applications.

\section*{Acknowledgment}
\label{sec:acknowledge}

This work is partly supported by the National Key Research and Development Plan (2024YFB3309302).

{
    \small
    \bibliographystyle{ieeenat_fullname}
    \bibliography{main}
}


\clearpage
\setcounter{table}{0}
\setcounter{figure}{0}
\setcounter{section}{0}
\renewcommand\thesection{\Alph{section}}
\renewcommand\thetable{\Alph{table}}
\renewcommand\thefigure{\Alph{figure}}
\maketitlesupplementary

\begin{figure*}[t]
  \centering
  \includegraphics[width=0.8\textwidth]{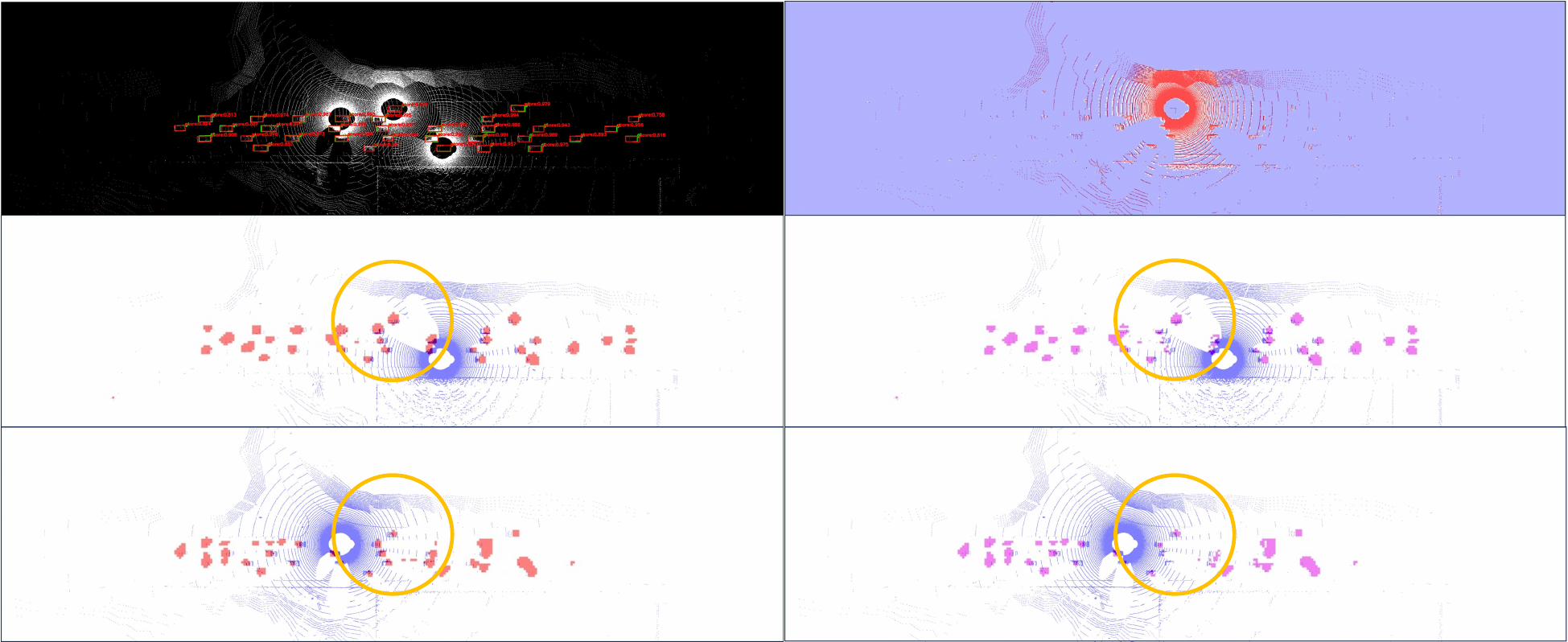}
  \caption{Visualization of supply and demand mask on the OPV2V~\cite{xu2022opv2v} dataset. The left side of the first row shows the collaborative perception detection results, while the right side shows the Ego agent's point cloud and its own demand mask (\textcolor[RGB]{160,160,254}{blue}). The left side of the second and third rows shows the point clouds of the two collaborating agents and their corresponding supply masks (\textcolor[RGB]{255,125,125}{red}), and the right side showing the supply-demand masks (\textcolor[RGB]{250,127,250}{pink}) combined with the Ego's demand mask.}
  \label{fig:vis2}
  \vspace{-5pt}
\end{figure*}

\begin{figure}[t]
  \centering
  \includegraphics[width=0.5\textwidth]{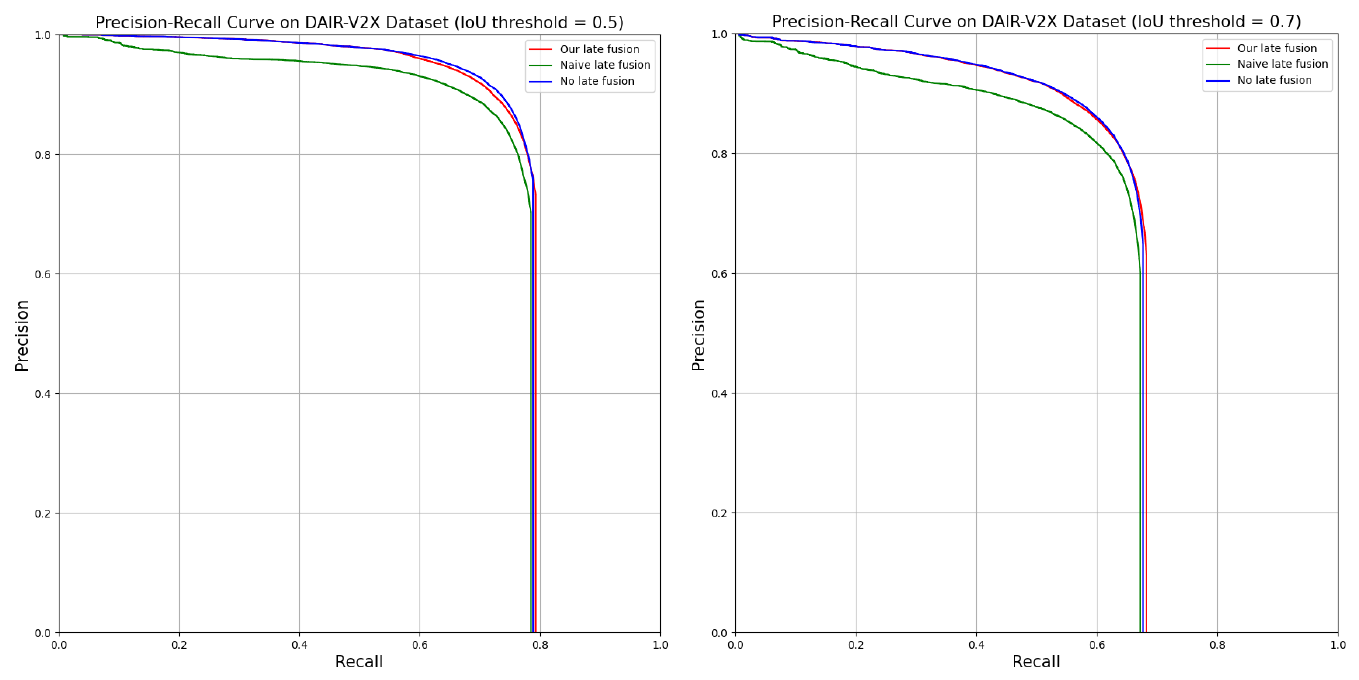}
  \caption{PR curves for different fusion schemes on DAIR-V2X~\cite{yu2022dair} dataset.}
  \label{fig:pr}
  \vspace{-5pt}
\end{figure}

\section{Dataset Details}
We evaluate the proposed \mymethodname~ against other methods on three different collaborative perception datasets (\textbf{OPV2V}~\cite{xu2022opv2v}, \textbf{V2XSim}~\cite{li2022v2x}, and \textbf{DAIR-V2X}~\cite{yu2022dair}) for LiDAR-based 3D object detection. There are more details about these datasets. 

OPV2V~\cite{xu2022opv2v} is a large-scale vehicle-to-vehicle (V2V) collaborative perception simulation dataset, obtained through the OpenCDA~\cite{xu2021opencda} and CARLA~\cite{dosovitskiy2017carla} simulators. It contains 11,464 frames of 3D radar point clouds and RGB images, along with 232,913 3D annotated bounding boxes. The training, validation, and test sets consist of 6,374, 2,170, and 1,980 frames, respectively. We set the perception range to $x \in [-140.8m, 140.8m], y \in [-38.4m, 38.4m]$.

V2XSim~\cite{li2022v2x} is a large-scale vehicle-to-everything (V2X) collaborative perception dataset, obtained using the SUMO~\cite{krajzewicz2012recent} and CARLA~\cite{dosovitskiy2017carla} simulators. It includes 10,000 frames of 3D radar point clouds and RGB images, with annotations for object detection, tracking, and semantic segmentation tasks. The training, validation, and test sets consist of 8,000, 1,000, and 1,000 frames, respectively. We set the perception range to $x \in [-32m, 32m], y \in [-32m, 32m]$.

DAIR-V2X~\cite{yu2022dair} is the first real-world vehicle-to-infrastructure (V2I) collaborative perception dataset, collected at an autonomous driving demonstration intersection in Beijing. It contains 38,845 frames of 3D radar point clouds and RGB images, with each frame containing data from a vehicle and a roadside infrastructure. This dataset is smaller, with 4,811 frames for the training set and 1,789 frames for the validation set. We set the perception range to $x \in [-140.8m, 140.8m], y \in [-40m, 40m]$. For the annotated data, we use the complete 360-degree annotations provided in~\cite{lu2023robust}

\section{More Visualization}
Fig.~\ref{fig:vis2} shows the regions selected by the supply-demand masks. From the first row, we can see that the demand mask covers the occluded areas near the Ego agent and regions with sparse point clouds at a distance, which are areas where Ego's perception is poor and require collaboration. The left side of the second and third rows shows the potential foreground regions selected by the supply masks of the two collaborating agents, which correspond closely to the ground truth in the detection results. The right side of the second and third rows shows the supply-demand masks of the two collaborating agents combined with the Ego agent’s demand mask. The regions circled in the figures show that the supply-demand mask selects fewer areas around the Ego agent compared to the left side, avoiding the selection of areas where the Ego agent has a good observation, thereby further reducing bandwidth. Due to the large perception range, the well-observed regions of the Ego agent account for only about $\sim$4\%, and the demand mask covers the majority of the area. However, because of occlusions, the demand mask contains fewer foreground regions. By combining supply masks, the bandwidth can be reduced by $\sim$10\% while maintaining detection accuracy.

Fig.~\ref{fig:pr} shows the PR (Precision-Recall) curves on the DAIR-V2X~\cite{yu2022dair} dataset for the scenarios without late fusion (intermediate fusion results), with naive late fusion, and with our confidence-aware late fusion. As can be seen, using naive late fusion introduces more suboptimal results, significantly lowering precision under low recall, and these results may overwrite the better detection results in the NMS (Non-Maximum Suppression) stage, causing a drop in overall recall. After using our confidence-aware late fusion, these suboptimal results are avoided, preventing a decrease in precision and improving recall, and it leads to an overall increase in AP (Average Precision).

\section{More Ablation}
Table~\ref{tab:ablation3} shows the differences in accuracy and bandwidth for various compression and fusion methods discussed in ``3.4 Message Compression and Fusion''. Since different compression and fusion methods affect the selection ratio of foreground regions, we removed the supply-demand selection module and transmit the complete BEV feature map, in order to more directly demonstrate the advantages of our method in terms of accuracy and bandwidth. From the table, it can be seen that the multi-scale fusion scheme has a significant advantage in accuracy compared to the traditional single-scale fusion scheme. Our multi-scale compression scheme significantly reduces bandwidth by compressing smaller intermediate features, and the fusion performed immediately after compression further improves accuracy.

\section{Discussion About the Latency}
Our \mymethodname~ requires the transmission of three messages between collaborative agents: demand masks, intermediate features, and detection results. For late fusion, we transmit the detection results of single cars rather than the detection results from intermediate fusion, as this can significantly reduce latency, allowing the detection results to be transmitted together with the intermediate features. In this way, \mymethodname~ only requires two rounds of inter-agent communication, which is still more than the common single-communication methods. In the following, we discuss the latency issues of \mymethodname.

1. \textbf{The total latency of \mymethodname~ is not necessarily higher.} (1) All collaborative methods require communication to transmit features. Since the communication volume is large, the transmission latency primarily depends on it (the larger the volume, the higher the latency). \mymethodname~ selects key regions according to the supply-demand relationship, which reduces the communication volume by more than 60\% while maintaining the highest accuracy, thereby significantly lowering the latency. (2) Another two communications with smaller volumes marginally contribute to the total latency as the system supports parallel processing. We record the latency of main components in Fig.~\ref{fig:framework} on OPV2V dataset: $t_{B} = 15\ \text{ms}$ (backbone), $t_{DG} = 1\ \text{ms}$ (demand generator), $t_{SG} = 3\ \text{ms}$ (supply generator). We roughly use $t_{C}=20\ \text{ms}$ as the latency for the two communications~\cite{grafling2010performance}. Since $t_{B} + t_{SG} = 18\ \text{ms} < t_{DG} + t_{C} = 21\ \text{ms}$, the transmission of demand masks only adds $3\ \text{ms}$ to the total latency. Similarly, transmitting detection results does not greatly increase the total latency.

\begin{figure}[]
  \centering
  \includegraphics[width=0.48\textwidth]{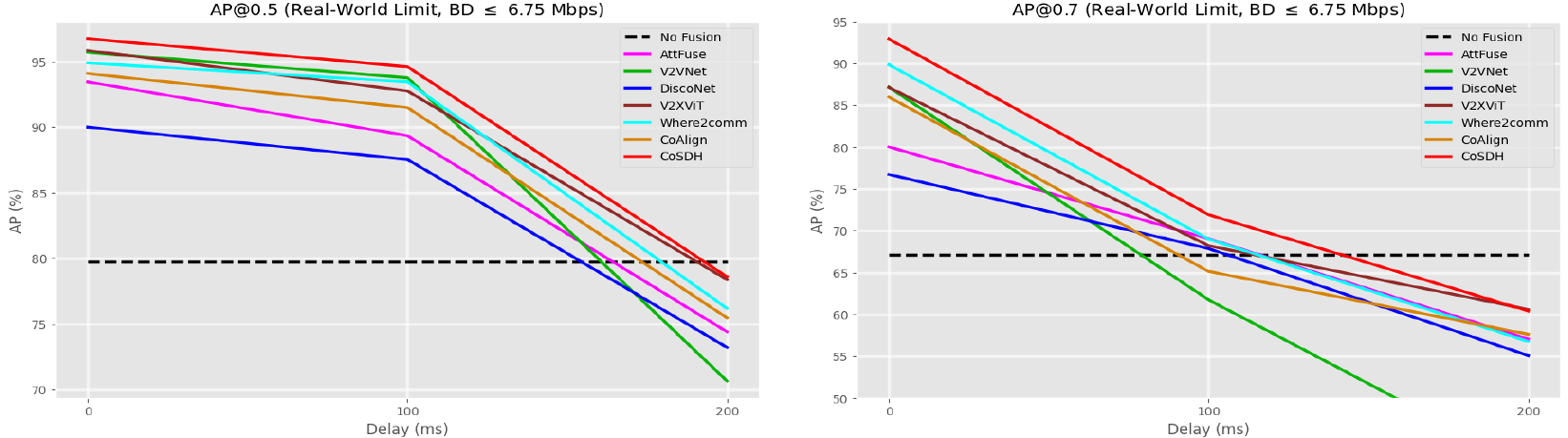}
  \caption{Robustness to latency on the OPV2V~\cite{xu2022opv2v} dataset.}
  \label{fig:delay}
  \vspace{-5pt}
\end{figure}

2. \textbf{\mymethodname~proves robust to the latency because of multi-scale fusion and collaborative region selection.} As shown in Fig.~\ref{fig:delay}, \mymethodname~ maintains the highest accuracy under a $100\ \text{ms}$ latency (the V2X communication based on IEEE 802.11p features a low latency, capable of keeping the latency within $100\ \text{ms}$). With $200\ \text{ms}$, almost all collaborative methods perform worse than ``No Fusion''. 

\begin{table}[]
\centering
\resizebox{0.48\textwidth}{!}{
\begin{tabular}{c|ccc|ccc}
\hline
       & \multicolumn{3}{c|}{Autoencoder}   & \multicolumn{3}{c}{Codebook~\cite{hu2024communication}}   \\ \cline{2-7} 
       & SF         & MF/SC     & MF/MC     & SF       & MF/SC    & MF/MC    \\ \hline
AP@0.5$\uparrow$ & 95.72\%    & 96.39\%   & 96.81\%   & 90.48\%  & 92.52\%  & 92.96\%  \\
AP@0.7$\uparrow$ & 91.18\%    & 92.31\%   & 93.00\%   & 84.32\%  & 87.40\%  & 88.17\%  \\
BD$\downarrow$     & 76.2 Mbps & 76.2 Mbps & 33.4 Mbps & 0.4 Mbps & 1.0 Mbps & 0.4 Mbps \\ \hline
\end{tabular}
}
\caption{Accuracy and bandwidth for different fusion and compression methods on the OPV2V~\cite{xu2022opv2v} dataset. ``SF'' represents single-scale fusion, ``MF/SC'' represents multi-scale fusion and single-scale compression, ``MF/MC'' represents multi-scale fusion and multi-scale compression. ``BD'' represents the bandwidth required for each collaborative agent, assuming the detection frequency is 10Hz.}
\label{tab:ablation3}
\end{table}

\end{document}